\documentclass[sigconf]{acmart}
\AtBeginDocument{%
  }

\copyrightyear{2025}
\acmYear{2025}
\setcopyright{acmlicensed}
\acmConference[MM '25]{Proceedings of the 33rd ACM International Conference on Multimedia}{October 27--31, 2025}{Dublin, Ireland}
\acmBooktitle{Proceedings of the 33rd ACM International Conference on Multimedia (MM '25), October 27--31, 2025, Dublin, Ireland}
\acmDOI{10.1145/3746027.3755326}
\acmISBN{979-8-4007-2035-2/2025/10}

\usepackage{multirow}
\usepackage{bbding}
\usepackage{pifont}
\usepackage{colortbl}
\usepackage{cleveref}
\usepackage{algorithm,algorithmicx}
\usepackage{algpseudocode}
\definecolor{mydarkblue}{rgb}{0,0.08,0.45}
\hypersetup{
	colorlinks=true,
	urlcolor=mydarkblue,
	citecolor=mydarkblue,
    linkcolor=mydarkblue
}
\usepackage{url}
\begin{document}

\title{NDM: A Noise-driven Detection and Mitigation Framework against Implicit Sexual Intentions in Text-to-Image Generation}

\author{Yitong Sun}
\authornote{Equal Contribution.}
\email{yt_sun@buaa.edu.cn}
\orcid{0009-0006-5294-2093}
\affiliation{%
  \institution{Institute of Artificial Intelligence, Beihang University}
  \city{Beijing}
  \country{China}}

\author{Yao Huang}
\authornotemark[1]
\email{y_huang@buaa.edu.cn}
\orcid{0000-0001-7978-2372}
\affiliation{%
  \institution{Institute of Artificial Intelligence, Beihang University}
  \city{Beijing}
  \country{China}}

\author{Ruochen Zhang}
\email{ruochen124@buaa.edu.cn}
\orcid{0009-0002-2514-6551}
\affiliation{%
  \institution{Institute of Artificial Intelligence, Beihang University}
  \city{Beijing}
  \country{China}}

\author{Huanran Chen}
\email{huanranchen@outlook.com}
\orcid{0000-0003-0847-9485}
\affiliation{%
  \institution{College of AI,\\ Tsinghua University}
  \city{Beijing}
  \country{China}}

\author{Shouwei Ruan}
\email{shouweiruan@buaa.edu.cn}
\orcid{0009-0007-0481-5855}
\affiliation{%
  \institution{Institute of Artificial Intelligence, Beihang University}
  \city{Beijing}
  \country{China}}

\author{Ranjie Duan}
\email{ranjie.drj@alibaba-inc.com}
\orcid{0009-0002-2261-4268}
\affiliation{%
  \institution{Security Group,\\ Alibaba Group}
  \city{Beijing}
  \country{China}}

\author{Xingxing Wei}
\email{xxwei@buaa.edu.cn}
\orcid{0000-0002-0778-8377}
\authornote{Corresponding Author.}
\affiliation{%
  \institution{Institute of Artificial Intelligence, Beihang University}
  \city{Beijing}
  \country{China}}

\renewcommand{\shortauthors}{Yitong Sun et al.}

\begin{abstract}
Despite the impressive generative capabilities of text-to-image (T2I) diffusion models, they remain vulnerable to generating inappropriate content, especially when confronted with implicit sexual prompts. Unlike explicit harmful prompts, these subtle cues, often disguised as seemingly benign terms, can unexpectedly trigger sexual content due to underlying model biases, raising significant ethical concerns. However, existing detection methods are primarily designed to identify explicit sexual content and therefore struggle to detect these implicit cues. Fine-tuning approaches, while effective to some extent, risk degrading the model’s generative quality, creating an undesirable trade-off. To address this, we propose NDM, the first noise-driven detection and mitigation framework, which could detect and mitigate implicit malicious intention in T2I generation while preserving the model’s original generative capabilities. Specifically, we introduce two key innovations: first, we leverage the separability of early-stage predicted noise to develop a noise-based detection method that could identify malicious content with high accuracy and efficiency; second, we propose a noise-enhanced adaptive negative guidance mechanism that could
optimize the initial noise by suppressing the prominent region's attention, thereby enhancing the effectiveness of adaptive negative guidance for sexual mitigation. Experimentally, we validate NDM on both natural and adversarial datasets, demonstrating its superior performance over existing SOTA methods, including SLD, UCE, and RECE, \textit{etc}. Code and resources are available at \textcolor{magenta}{\url{https://github.com/lorraine021/NDM}}.   
\end{abstract}

\ccsdesc[300]{Security and privacy~Human and societal aspects of security and privacy}
\keywords{Text-to-Image Generation, Implicit Sexual Intentions, Noise-driven Detection and Mitigation}

\begin{teaserfigure}
    \centering
    \includegraphics[width=0.85\linewidth]{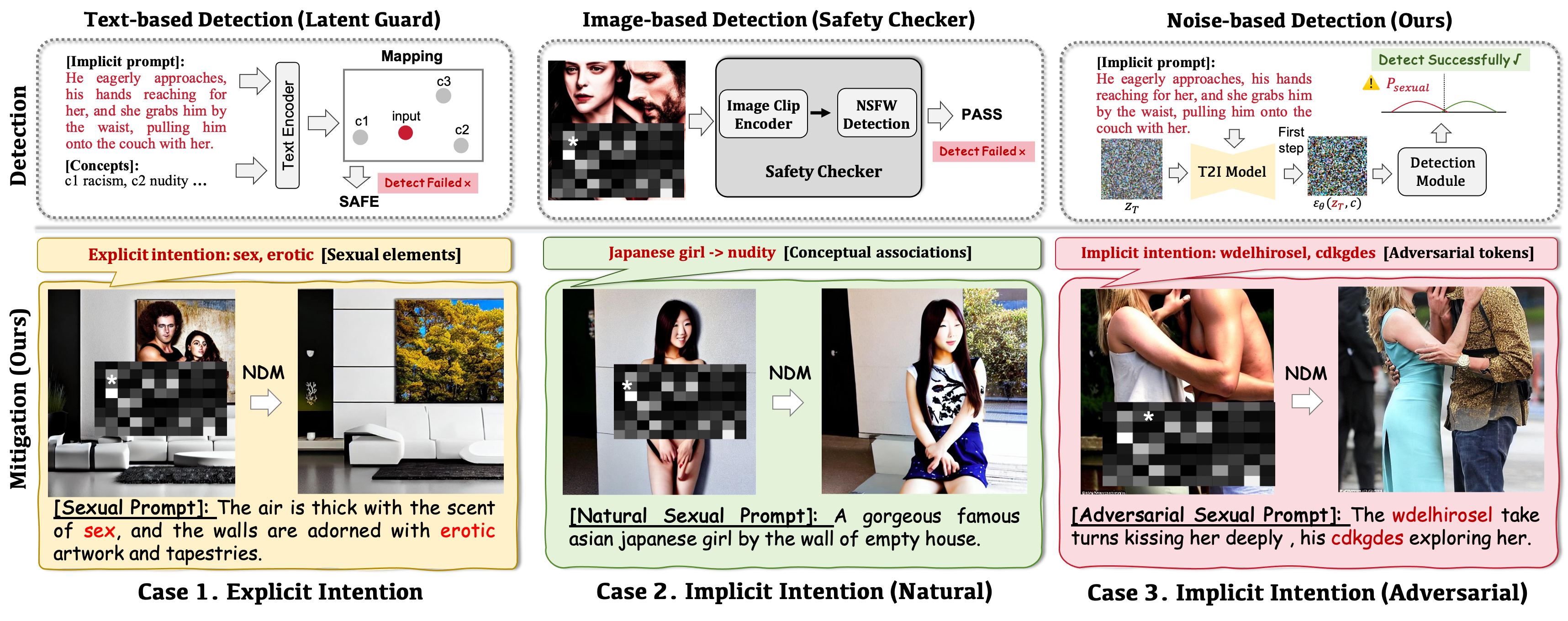}
    \caption{\textit{Up:} Detection challenges in existent methods. Text-based detections fail to detect implicit sexual intent due to reliance on prompt encoding and harmful concept comparison. Image-based methods, which map to the CLIP space, are hindered by interference and the need for fully generated images. In contrast, our NDM leverages early-stage predicted noise, achieving superior efficiency and precision in detecting harmful content. \textit{Bottom:} An illustration of our NDM's successful mitigation across various settings: explicit intention, implicit intention (natural and adversarial), showcasing its broad effectiveness. 
}
    \label{fig:cover}
    \Description{cover image}
\end{teaserfigure}


\maketitle

\section{Introduction}
Recent advances in diffusion models~\cite{nichol2021glide, rombach2022high, Midjourney} have propelled text-to-image (T2I) techniques to achieve remarkable performance in synthesizing photorealistic images from textual prompts, driving widespread adoption in diverse domains, including digital art creation~\cite{wang2024diffusion}, advertising product visualization~\cite{saharia2022photorealistic}, and medical image synthesis~\cite{chambon2022roentgan, aversa2023diffinfinite}. However, their powerful generative capabilities also present significant risks. Specifically, they can be exploited to produce inappropriate content, especially pornography \cite{schramowski2023safe, qu2023unsafe, yang2024sneakyprompt} like explicit imagery that mimics real individuals or even illegal material like child exploitation, raising serious ethical concerns.

To mitigate the ethical challenges posed by T2I techniques~\cite{hagendorff2024mapping,bird2023typology}, prior research has explored a range of strategies. These efforts can be broadly categorized into two main categories: \textit{Model-intrinsic methods} and \textit{Model-extrinsic methods}. \textbf{\textit{Model-intrinsic methods}} generally modifies the model's internal parameters. Techniques such as fine-tuning CLIP weights~\cite{poppi2024safe}, concept unlearning~\cite{gandikota2023erasing, huang2024receler}, and parameter editing~\cite{gandikota2024unified} are employed to suppress explicit content. While these methods are effective in mitigating known undesirable outputs, they often suffer a significant trade-off, as they degrade overall generation performance.

\begin{figure*}[!t]
    \centering
    \includegraphics[width=0.98\linewidth]{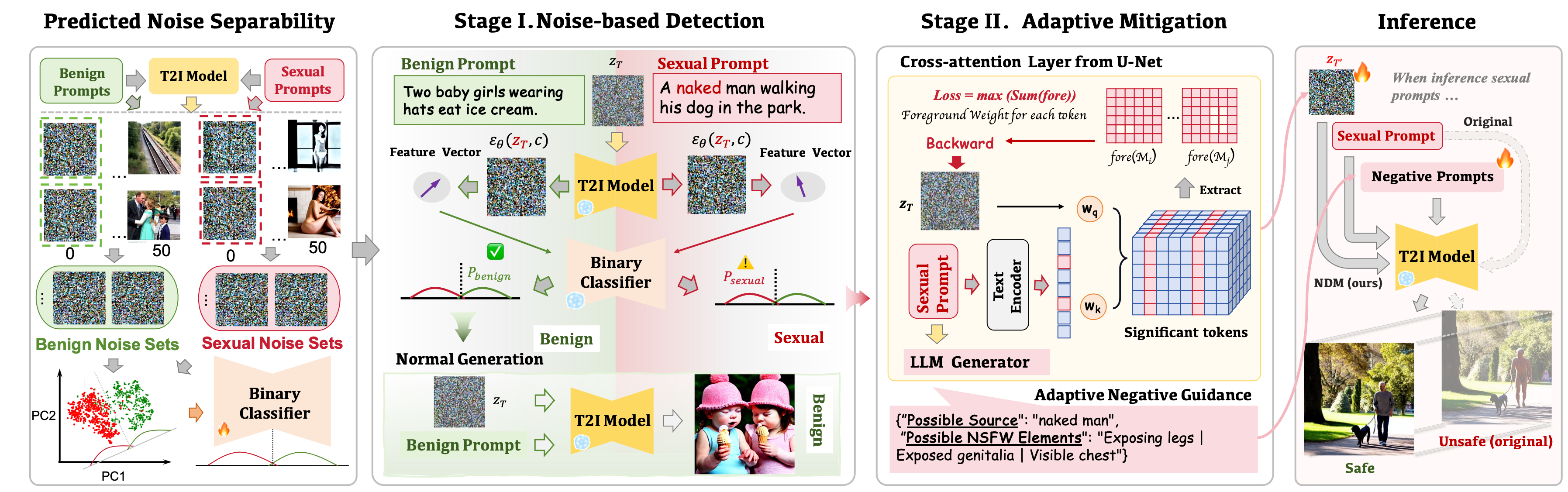}
    \caption{Overview of the NDM framework. \textit{Stage I}: Noise-based Detection utilizes predicted noise separability to classify benign and sexual prompts. \textit{Stage II}:  When sexual prompts are detected, adaptive mitigation begins by optimizing the initial noise through suppressing significant foreground regions in the cross-attention map. This is followed by combining the optimized noise with adaptive negative prompts generated by an LLM, tailored to the input prompts for more effective sexual mitigation.} 
    \Description{framework}
    \label{fig:framework}
\end{figure*}

In contrast, \textbf{\textit{Model-extrinsic methods}} focus on detection and mitigation to block sensitive content, which could better preserve the performance on regular tasks without internal modifications. 
Some methods use external safeguards, such as plug-and-play filters, to detect inappropriate outputs via textual cues~\cite{liu2024latent} or generated imagery~\cite{safetychecker}; some methods steer generation in an opposing or harmless direction, like steering prompt embeddings away from harmful subspaces~\cite{yoon2024safree} or guiding away from unsafe prompts~\cite{rombach2022high}. However, they still struggle with implicit malicious intent from both subtle conceptual associations in training data and adversarial inputs~\cite{tsai2024ring, chin2024prompting4debugging, yang2024sneakyprompt, zhang2024generate}. As depicted in \Cref{fig:cover}, benign phrases like "Japanese girl" may trigger harmful content like "nudity" due to latent data biases, and optimized adversarial tokens (e.g., ``wdehirosel'', ``cdkgdes'') can manipulate behavior without triggering conventional filters, highlighting a critical unresolved gap.

Thus, this paper aims to ensure safer text-to-image generation by \textit{inheriting the detection-and-mitigation framework's advantage of alleviating trade-offs while crucially addressing the issue of implicit malicious intention}. Naturally, two key challenges emerge: (1) \textbf{how to improve the detection of implicit malicious intention early}? Existing text-based detectors primarily focus on explicit harmful content but struggle to capture subtle, implicit malicious intent, often hidden within seemingly benign prompts. Image-based detection methods require an image's full generation before assessing, introducing significant delays. Therefore, more efficient and accurate detection techniques are needed. 
(2) \textbf{how to effectively mitigate implicit malicious intention during generation}? Existing methods focus on steering away from predefined harmful subspaces, but this paradigm fails to handle diverse implicit malicious content arising from complex and varied prompts. Also, simple negative guidance alone may not be sufficient to prevent certain significant malicious outputs. To overcome this, we need a dynamic, context-aware, and enhanced mechanism that can adapt in real-time to various implicit sexual prompts, enabling flexible and effective mitigation throughout the generation.

To meet the above goal, we innovatively propose \textbf{NDM}, a \textbf{\underline{N}}oise-driven \textbf{\underline{D}}etection and \textbf{\underline{M}}itigation framework, which could address implicit malicious intention in text-to-image generation. To be specific, for the first challenge, we draw inspiration from a key observation in the diffusion process: the denoising procedure is inherently coarse-to-fine, with early steps defining the main structure of the image and later steps refining the details. Thus, the early-stage predicted noise, especially from the first few denoising steps, exhibits significant separability between normal and sexually explicit images (as depicted in \Cref{fig:noise}). This insight leads us to utilize the early-stage noises to train a classifier for detecting implicit malicious intention, which greatly improves the accuracy of detection. Moreover, since this method performs detection based on the noise from the initial denoising steps, it incurs virtually no additional computational cost compared to traditional ones~\cite{safetychecker, liu2024latent}.

For the second challenge, we propose a noise-enhanced adaptive negative guidance. Instead of using predefined, static negative prompts like “nudity”, we dynamically generate negative prompts tailored to the inputs using a large language model (LLM) to better capture prompt nuances and identify which harmful elements to avoid. Furthermore, inspired by the significant influence of initial noise on generated content~\cite{guo2024initno, qi2024not, xu2025good, ban2024crystal}, we innovatively explore initial noise's effects on safety by analyzing the frequency of nudity appearance from different initial noises, and we gain an insightful observation: Different initial noises vary in explicit content generation, which means a better choice can reduce unethical imagery. Thus, we further perform an initial noise optimization by suppressing prominent malicious attention, providing a safer starting point for negative guidance.
Main contributions are as follows:

 \ding{182} \textbf{We introduce NDM, the first noise-driven detection and mitigation framework,} which could ensure safer image generation while preserving the model's general generative capabilities.

 \ding{183} \textbf{We uncover two key insights into noises for safe text-to-image generation:} the separability of early-stage predicted noises (\textit{allowing for efficient detection}) and the significant impact of initial noises on sexual content generation (\textit{leading to a more effective noise-enhanced adaptive negative guidance for mitigation}).
    
\ding{184} \textbf{We comprehensively evaluate our method on both natural implicit and adversarial datasets for sexual content detection and mitigation}. Experimental results verify the superior effectiveness of our NDM for different implicit sexual prompts when compared with other SOTA methods, \textit{e.g.}, SLD, UCE, and RECE, \textit{etc}.

\section{Related Works}
\subsection{Ethical Risks with T2I Generation}
As T2I generation models advance, several ethical risks~\cite{zhang2024multitrust, xu2025mmdt, huang2025perception} also emerge, particularly regarding the generation of sexual content.  To systematically assess this, Schramowski \textit{et al.} propose the I2P dataset~\cite{schramowski2023safe}, a collection of malicious prompts designed to evaluate the generation of inappropriate imagery. Their work reveals that open-source latent diffusion models, such as Stable Diffusion~\cite{rombach2022high}, continue to struggle with ensuring safe content generation. Among these, sexual content, which arises from implicit associations and underlying concepts rather than explicit statements, represents one of the most significant threats. Beyond this, some other studies have demonstrated that diffusion models are also vulnerable to sexual content caused by implicit adversarial manipulation.
For example, Prompting4Debugging~\cite{chin2024prompting4debugging} and Ring-a-bell~\cite{tsai2024ring} employ prompt engineering techniques to generate seemingly benign inputs but could lead to harmful outputs, akin to jailbreaks in LLMs~\cite{huang2025breaking,zeng2024johnny}. Similarly, SneakyPrompt~\cite{yang2024sneakyprompt} uses reinforcement learning to discover adversarial prompts that bypass safety filters while preserving harmful semantics. MMA-Diffusion~\cite{yang2024mma} further exploits both textual and visual inputs to evade the model’s safeguards. These studies underscore the pressing need for more robust countermeasures to address the risks posed by such implicit sexual prompts. 

\begin{figure*}[!t]
    \centering
    \includegraphics[width=0.97\linewidth]{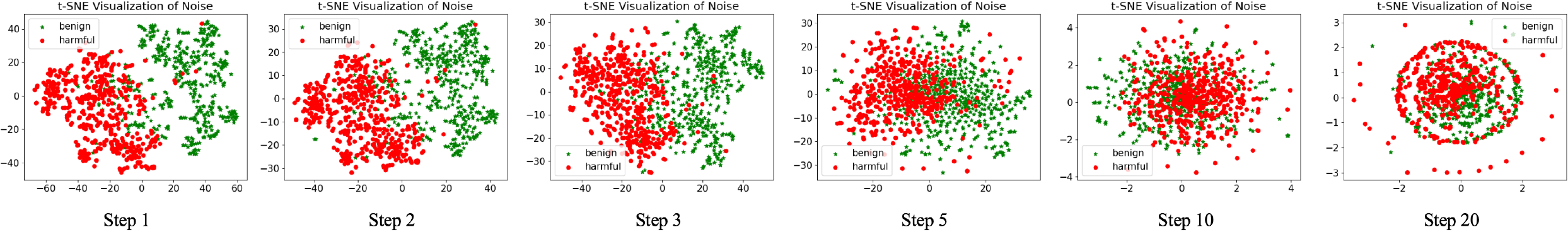}
        \caption{Visualized separability of predicted noises at different timesteps for benign and sexual generations using t-SNE.} 
        \Description{noise distribution}
    \label{fig:noise}
\end{figure*}

\subsection{Defense Against Malicious Generation}
Significant efforts have been made to explore defense strategies, which can be broadly divided into model-intrinsic methods and model-extrinsic methods. 
Model-intrinsic methods tend to modify internal parameters of the model to suppress harmful outputs. Unlearning approaches like ESD~\cite{gandikota2023erasing} and Receler~\cite{huang2024receler} are the most classical ones, which focus on denoising by aligning predicted noises with negatively guided distributions or steering outputs toward neutral targets based on fine-tuning. Similarly, Safe-CLIP~\cite{poppi2024safe} fine-tunes the text encoder's weights in CLIP to reduce sensitivity to harmful inputs. Model-editing techniques, such as UCE~\cite{gandikota2024unified} and RECE~\cite{gong2024reliable}, modify specific layers like cross-attention weights to achieve efficient suppression of harmful content. Also inspired by safety alignment techniques in LLMs~\cite{zhang2025stair, zhang2025realsafe}, some methods~\cite{ruan2025towards, liu2024safetydpo} introduce safety constraints in DPO-based training.
Yet, these model-intrinsic methods still face challenges with degradation in non-malicious generation. 
On the other hand, model-extrinsic methods focus on external interventions to block harmful content without altering internal parameters. For instance, methods like Latent Guard~\cite{liu2024latent} and Stable Diffusion’s safety checker~\cite{safetychecker} detect harmful concepts within the model’s latent space or generated images and then intervene in the output. Additionally, techniques such as SLD~\cite{schramowski2023safe} and Safree~\cite{yoon2024safree} steer prompt embeddings away from harmful subspaces, effectively balancing toxicity filtering with concept preservation. However, these methods still struggle with prompts with implicit malicious intention, arising from subtle associations or adversarial inputs that remain undetected by conventional detectors. 
Therefore, our NDM aims to not only ensure safe generation against implicit sexual prompts but also preserve the performance on non-malicious input prompts in the meantime.

\section{Methodology}
In this section, we will detail our NDM framework, as shown in \Cref{fig:framework}. NDM addresses the issue of handling implicit sexual prompts through a novel noise-based framework. Specifically, we will discuss how noise can be leveraged for high-accuracy and efficient detection (\Cref{sec:noise_detection}), and how adaptive negative guidance and optimizing the initial noise further enhance mitigation (\Cref{sec:noise_mitigation}). We first introduce the necessary background in \Cref{sec:pre}.

\subsection{Preliminaries}
\label{sec:pre}
\textbf{T2I Diffusion Models:} Text-to-image diffusion models, especially latent diffusion models, have demonstrated remarkable performance by generating high-quality images from textual prompts. These models generate images by iteratively refining a latent representation from random noise, guided by the input prompts. Specifically, 
the process begins with a random Gaussian noise sampled from a standard normal distribution $z_T \sim \mathcal{N}(0, I)$, where $ z_T $ represents the initial latent variable at time step $ T $. Then, at each subsequent time step $ t $, the model uses a conditional text embedding $ c $, encoded by a CLIP model, to predict the noise $ \epsilon_\theta(z_t, c) $. The denoising operation at each step progressively refines the latent representation by adjusting $ z_{t-1} $, under classifier-free guidance~\cite{ho2022classifier}. This guidance combines both an unconditional prediction $ \epsilon_\theta(z_t, \emptyset) $ (with no text input) and the conditional prediction $ \epsilon_\theta(z_t, c) $ (based on the text embedding $ c $), effectively balancing creativity and fidelity in the generated image, formulated as follows:
\begin{equation}
z_{t-1} = \epsilon_\theta(z_t, \emptyset) + \gamma \cdot \left( \epsilon_\theta(z_t, c) - \epsilon_\theta(z_t, \emptyset) \right),
\label{eq:1}
\end{equation}
where $ \theta $ denotes the parameters of the diffusion model, and $ \gamma $ is a scalar guidance scale controlling the strength of the classifier-free guidance. At the end of denoising, the model decodes the last latent representation $ z_0 $ back into pixel space to obtain the image $I$.

\noindent\textbf{Cross-Attention in U-Net:} In the denoising process, the U-Net architecture plays a central role, particularly through its cross-attention layers, which integrate the text embedding $ c $. These cross-attention layers allow the model to focus on specific regions of the latent representation that are influenced by the text embedding. The cross-attention mechanism is described by the following formula:
\begin{equation}
M = \text{softmax}\left( \frac{Q K^T}{\sqrt{d}} \right),
\end{equation}
where $ M $ is the cross-attention map, and $ M_i $ denotes the attention map of the $ i $-th token. Specifically, $ M_i[x, y] $ represents the attention weight at spatial coordinates $ [x, y] $ for the $ i $-th token.

In NDM, we also focus on the widely used latent diffusion models. The details of our noise-based detection and mitigation framework are presented in the following sections.

\subsection{Noise-Based Sexual Detection}
\label{sec:noise_detection}

Existing text-based detection methods struggle with implicit sexual content, particularly when prompts lack explicit cues (\textit{e.g.}, “a woman in a bedroom”). This failure arises because text-based detection methods cannot capture the correlations between seemingly benign prompts and the harmful visual patterns associated with them. On the other hand, image-based detection methods, such as safety checker~\cite{safetychecker} require the full generation of an image $I$ before assessing potential harm, which introduces great inefficiency. 

To address these challenges, inspired by image-based detection methods that use fundamental visual semantics to detect implicit sexual prompts, we seek to explore whether the predicted noise during the denoising process can serve a similar function, which could simultaneously reduce computational cost by allowing earlier detection. Since diffusion models refine images from coarse to fine details~\cite{xu2025good}, the noise at earlier timesteps may already capture critical features that distinguish sexual content from benign content.

\textbf{Early-stage Predicted Noise Separability:}  To verify the feasibility of this hypothesis, we analyze the predicted noise at different timesteps during the denoising process. Specifically, we select 500 sexual prompts from the I2P dataset~\cite{schramowski2023safe} and 500 benign prompts from the COCO-30k dataset~\cite{lin2014microsoft}. Using Stable Diffusion v1.4~\cite{rombach2022high}, we extract the predicted noise sets $ \{\epsilon_b\}_t $ and $ \{\epsilon_s\}_t $ at various stages of the denoising process. We then apply t-SNE~\cite{van2008visualizing} to visualize the separability of the noise distributions across different timesteps, aiming to determine whether distinct separability between harmful and benign content emerges early in the denoising process.

 As shown in \Cref{fig:noise}, the early-stage predicted noise already exhibits distinct patterns that could differentiate harmful content from benign content. Moreover, these differences are more pronounced in the initial few steps and gradually diminish as the denoising process progresses, which suggests that the influence of the input prompt $c$ is more significant in the early stages of the denoising process. Thus, it is reasonable to train a classifier using early-stage predicted noise to detect sexual content, whether explicit or implicit.

\textbf{Detection Model Training:} 
The objective of training is to construct a binary classifier $\mathcal{F}: X_{\text{input}} \to \{0, 1\}$:

\begin{equation}
\mathcal{F}(X_{\text{input}}) =
\begin{cases} 
1, & \text{if } \mathcal{F}(X_{\text{input}}) \in \mathcal{P}_{\text{sexual}} \\
0, & \text{otherwise}
\end{cases},
\label{eq:f_objective}
\end{equation}
where $F(X_{\text{input}})=1$ indicates that the input is likely to steer the model to generate sexual content and $\mathcal{F}(X_{\text{input}})=0$ suggests safe.

Then, based on the above insights, our classifier trains on the first-step predicted noise $\epsilon_1$ from the diffusion U-Net. This procedure consists of sequential parts: We first adopt PCA~\cite{abdi2010principal} to conduct noise decomposition and capture dominant patterns, then we use LDA~\cite{fisher1936use} to maximize the discrimination of the two groups, and finally we employ a classical yet effective classification model, SVM~\cite{cortes1995support} to fit the processed feature vectors and build the decision boundary, which could be expressed as:
\begin{equation}
   \mathcal{F}(X_{\text{input}}) = \text{sign} \left( \mathbf{w}^\top \mathbf{W}_{\text{lda}}^\top \mathbf{W}_{\text{pca}}^\top (X_{\text{input}} - \boldsymbol{\mu}) + b \right),
    \label{eq:f}
\end{equation}
where $\mathbf{W}_{\text{pca}} \in \mathbb{R}^{d \times k}$ is the projection matrix resulting from PCA, with $k = 2$, $\mathbf{W}_{\text{lda}} \in \mathbb{R}^{k \times m}$ is the projection matrix obtained from LDA, with $m = 1$, $\mathbf{w} \in \mathbb{R}^m$ is the weight vector of the SVM classifier, $\boldsymbol{\mu}$ denotes the mean of the training data, and $b$ represents the bias term.
Overall, by training on early-stage predicted noise from both sexual and benign inputs, the classifier is able to effectively differentiate between the two classes, resulting in high accuracy and robust generalization for detecting sexual inputs.

\subsection{Noise-Enhanced Adaptive Mitigation}
\label{sec:noise_mitigation}
\textbf{Adaptive Negative Guidance: }
After identifying sexual inputs, whether explicit or implicit, the next step is to replace exposed pornographic elements with appropriate alternatives, such as covering nudity with clothing. For instance, if the original image output $I$ corresponds to the scene “a person with a bare torso standing on the beach,” the processed output $I'$ should depict “a person wearing clothes standing on the beach.” This replacement ensures that the image content aligns with social ethics by covering inappropriate exposure while preserving other visual elements of the original scene. We follow~\cite{rombach2022high}, which enables safe imagery using negative prompt guidance to protect against sexual elements. In this context, the denoising step modifies the unconditional predicted noise $\epsilon_{\theta}(z_t, \emptyset)$ in \Cref{eq:1} to the negative counterpart, with the denoising process defined as:
\begin{equation}
    z_{t-1} = \epsilon_\theta(z_t, c_{\text{neg}}) + \gamma * (\epsilon_\theta(z_t, c) - \epsilon_\theta(z_t, c_{\text{neg}})),
    \label{eq:z_neg}
\end{equation}
where $c_{\text{neg}}$ represents the text embedding of the negative prompt. By following this denoising process step-by-step, the final imagery $I$ effectively avoids the undesired concepts introduced by $c_{\text{neg}}$.

However, previous works often use fixed abstract concepts like "nudity" as the target $c_{\text{neg}}$ for all cases. This paradigm faces challenges, as a generic "nudity" prompt is not always sufficient. To address this, we propose an adaptive negative guidance mechanism, leveraging the powerful language comprehension capabilities of a large language model (LLM) ($\mathcal{G}$). Our method analyzes nouns, verbs, and adjectives of the input prompts, which are key linguistic components in determining the content of a prompt. Specifically, nouns are crucial for identifying subjects that could be associated with inappropriate content; verbs describe actions or behaviors that might introduce explicit or suggestive themes; and adjectives are important for refining the properties of these nouns and verbs, potentially highlighting sexually explicit characteristics. After analyzing these linguistic features, the LLM predicts and maps potential indirect expressions to expressions with more specific features of the possible sexual imagery, which allows the diffusion model to more clearly understand the specific visual elements it should avoid. We define this process as follows:
\begin{equation}
    c_{\text{neg}} = \text{CLIP}(\mathcal{G}(p; \mathcal{V}_{\text{noun}}, \mathcal{V}_{\text{verb}}, \mathcal{V}_{\text{adj}})),
    \label{eq:c_neg}
\end{equation}
where $ \mathcal{G} $ represents the LLM, which processes the input prompt $p$ with respect to its components in the noun, verb, and adjective vocabularies: $ \mathcal{V}_{\text{noun}}, \mathcal{V}_{\text{verb}}, \mathcal{V}_{\text{adj}} $. The output of $\mathcal{G}$ is then mapped to the negative prompt embedding $c_{\text{neg}}$ using CLIP, guiding the model to avoid generating sexual content.  However, experimental results show that such an adaptive negative guidance alone remains insufficient at times. Thus, to achieve more effective mitigation, we need to explore joint efforts.

\begin{figure}[!t]
    \centering
    \includegraphics[width=\linewidth]{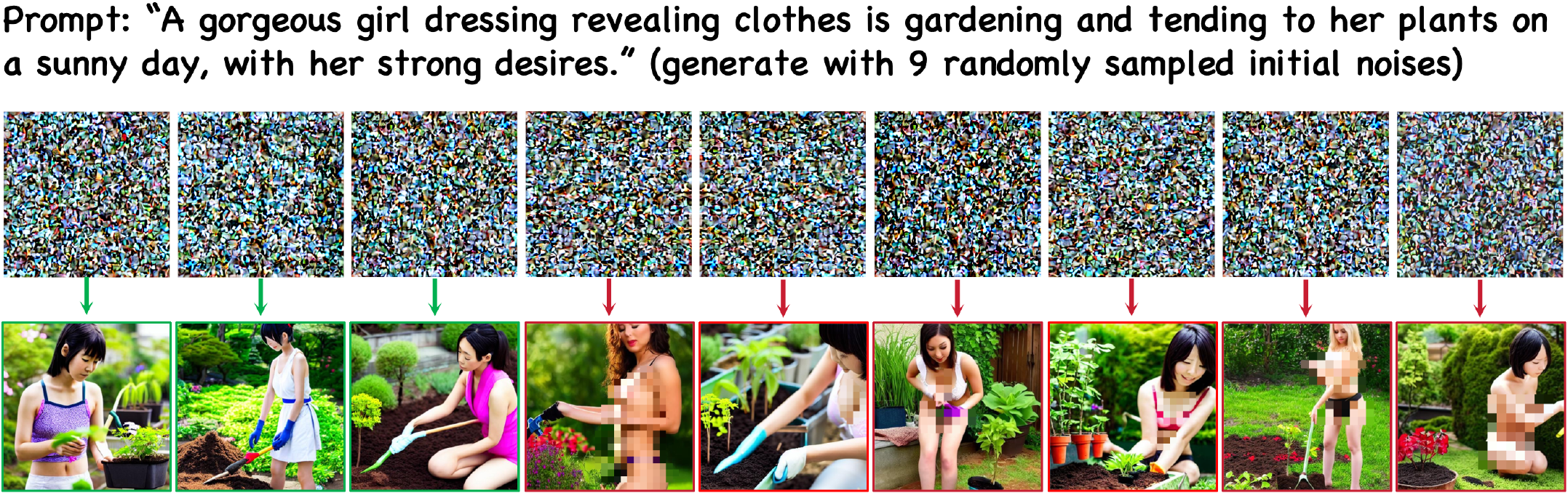}
        \caption{Generation results of the same prompt under various sampled initial noises. The cases framed in $\textcolor{teal}{\text{green}}$ are safe, while the cases framed in $\textcolor{red}{\text{red}}$ are sexual.} 
        \Description{noise distribution}
    \label{fig:initial}
\end{figure}

\textbf{Initial Noise Optimization:} 
Prior works~\cite{guo2024initno, qi2024not, ban2024crystal} have explored the impacts of initial noise on diffusion models' generation quality, highlighting the significance of the initial noise $z_T$. Xu \textit{et al.}~\cite{xu2025good} further validate this by swapping seeds at different stages during reverse diffusion. Their results show that the initial noise strongly affects the generated content, while subsequent noise adjustments have minimal effects.

Building on the above findings, we aim to extend this by exploring a causal relationship: how initial noise $z_T$ impacts sexual element expressions in the generated image $I$. To explore this, we randomly sample initial noises for generation using sexual prompts. As shown in \Cref{fig:initial}, we observe significant variation in how different initial noises trigger pornographic elements under the same prompt. This confirms that \textit{the initial noise indeed plays a crucial role in shaping the manifestation of pornographic elements}. 
Based on this, we aim to design a method that can optimize the initial noise for a better starting point of the adaptive negative guidance.

But how can we optimize the initial noise? To address this, we first analyze the attention weights of different tokens in the input prompt. Specifically, we follow~\cite{guo2024initno} to extract the cross-attention maps $M_i, i=1, \cdots, n$ of tokens and identify the maximum attention value $ \max(M_i), i=1, \cdots, n$. We observe that only one or two tokens have attention weights exceeding 0.1, indicating a skewed attention distribution. This suggests that certain key tokens disproportionately influence the model's behavior, making it challenging to intervene or modify the model’s decisions due to their absolute dominance. Actually, in sexual image generation, the most prominent tokens often correspond to the explicit sexual elements. Therefore, a natural approach for optimizing the initial noise is to reduce the attention given to these dominant tokens.
 
To achieve finer-grained optimization, we first skip stopwords and other nonsense words, focusing on tokens that carry meaning within the input. By isolating these meaningful tokens, we can better analyze their individual contributions. A direct objective is to manipulate $max(M_i)$. However, simply suppressing the maximum value may not be sufficient, as the attention map of the dominant token may contain other significant values that, when aggregated, still contribute considerable weight. To address this issue, we propose a new attention quantification metric $ Sum_i $, which considers the sum of the attention weights in the foreground region \( \Omega_i \) associated with token $i$. This metric reflects both the size of the foreground and the strength of the control exerted by each token. Specifically, we first define the foreground region \( \Omega_i \) of token $i$ as:
\begin{equation}
    \Omega_i = \{\Omega_i[x, y] \mid \Omega_i[x, y] = 1(M_i[x, y] > \beta), \forall (x, y)\},
    \label{eq:omega_i}
\end{equation}
where \( \Omega_i \) is a binary mask that contains coordinates whose attention weight exceeds a threshold \( \beta \), which is adaptively computed using Otsu's method \cite{otsu1975threshold}. Then, we could calculate the sum of the original foreground weights in token $i$'s cross-attention map:
\begin{equation}
     Sum_i = \sum_{(x, y) \in \Omega_i} M_i[x, y], \quad i=1,2,...,n.
    \label{eq:sd1}
\end{equation}
 
Finally, the optimization objective focuses on the largest value among all $Sum_i$ instead of $M_i$, which actually provides a stronger optimization signal. The loss function could be computed as follows:
\begin{equation}
     \mathcal{L}_{cross} = \max_{i} (Sum_i), \quad i=1,2,...,n,
    \label{eq:sd2}
\end{equation}
where the loss $\mathcal{L}_{cross}$ emphasizes the regional influence of the most dominant token. The iteration process continues until the loss decreases to a fraction of its initial value, specifically $\mathcal{L}_{cross} \leq \alpha \cdot \mathcal{L}_{init}$, and $\alpha \in (0, 1)$ is a hyperparameter that controls the extent of semantic intensity reduction. By introducing this stopping criterion, we ensure that the attention dominance is gradually mitigated while maintaining a controllable level of semantic weakening.

\section{Experiments}
\subsection{Experimental Settings}
\noindent\textbf{Evaluation Datasets:}
We first evaluate on five sexual datasets. These include I2P dataset~\cite{schramowski2023safe} (931 prompts for sexual generation) and adversarial prompts from the following methods: Sneaky Prompt~\cite{yang2024sneakyprompt} (SP(N) and SP(P) for natural and pseudo conditions, 200 each), Ring-A-Bell~\cite{tsai2024ring} (79 prompts) and MMA-Diffusion~\cite{yang2024mma} (1000 prompts). 
Additionally, we include 1000 prompts from COCO-30k~\cite{lin2014microsoft} for benign generation evaluation.

\noindent\textbf{Compared Baselines:}
We totally evaluate two settings: detection-then-refusal and detection-then-mitigation. For the former, we consider several detection methods, including Text Blacklist Checking~\cite{Eileen2023}, CLIP Score~\cite{hessel2021clipscore}, Latent Guard~\cite{liu2024latent}, and safety checker~\cite{safetychecker}. 
BERT Score is excluded due to its ineffectiveness in detecting semantic errors as discussed in~\cite{zhang2019bertscore}.
For the latter, we compare against a comprehensive set of baselines (training-based and training-free): SLD~\cite{rombach2022high} (weak/medium/strong/max), UCE~\cite{gandikota2024unified}, RECE~\cite{gong2024reliable}, ESD~\cite{gandikota2023erasing}, Safree~\cite{yoon2024safree}, Safe-CLIP~\cite{poppi2024safe}, and Stable Diffusion v2.1~\cite{rombach2022high}, which is trained on filtered LAION~\cite{schuhmann2022laion} for safety.
We conduct three trials for each and report the best results. More details are listed in \textit{Appendix}.

\noindent\textbf{Evaluation Metrics:}
To assess whether and to what extent the outputs contain sexual elements, we employ the NudeNet classifier~\cite{nudenet2019} with a threshold of 0.45 to calculate the attack success rate (ASR), and the NudeNet detector~\cite{nudenet2019} also with a threshold of 0.45 to compute the nudity removal rate (NRR) following the ESD approach. For the detector, we focus on exposed components, including "exposed female/male breast", "exposed female/male genitalia", and "exposed buttocks". Additionally, we also evaluate non-sexual semantics preservation using CLIP Score~\cite{hessel2021clipscore} and FID Score~\cite{heusel2017gans}, with higher CLIP and lower FID indicating better alignment with prompts and greater similarity to SD-v1.4-base images, respectively.

\noindent\textbf{Implementation Details:}
Following prior methods~\cite{gandikota2023erasing, guo2024initno}, we use Stable Diffusion v1.4~\cite{rombach2022high} as the base model for testing, setting the guidance scale \( \gamma \) to the fixed setting of 7.5 and the denoising steps \( T \) to 50. For detection, we increase the guidance scale \( \gamma \) to 12.5 to achieve stronger semantic injection. In optimizing the initial noise, we set the threshold \( \alpha \) to 0.7 and limit the maximum number of optimization iterations to 30. For the detection module, we train the classifier with the first-step predicted noises from 500 sexual prompts from the I2P and 500 benign prompts from COCO-30k. For computational resources, we utilize an NVIDIA GeForce RTX 3090 with 24GB GPU memory.

\subsection{Effectiveness of Sexual Detection}
To validate our noise-based detection, we compare it with prior text-based and image-based detection approaches. As shown in \Cref{tab:detect}, most text-based methods like Blacklist and LatentGuard struggle to detect sexual intent due to missed trigger words, greatly compromising the detection reliability. While CLIP Score, as an image-based method, performs relatively well, but is highly sensitive to threshold choice, leading to inconsistent performance, especially on borderline cases. Image-based methods like the safety checker are somewhat effective but suffer from noticeable missed detections. In contrast, our detection module proves more robust, accurate, and consistent across natural and adversarial scenarios.
Moreover, it excels in detection speed with only \textasciitilde 0.95 s/sample, significantly outperforming image-based methods while matching text-based ones, making it a highly efficient solution for real-time sexual content moderation without sacrificing accuracy.

\begin{table}[!h]
    \centering
    \caption{Comparison with other detection methods.}
    \label{tab:detect}
    \resizebox{\linewidth}{!}{
    \begin{tabular}{l|cccc|c|c}
    \toprule[1.5pt]
      \textbf{Method}  & \textbf{I2P} & \textbf{SP(N)} & \textbf{SP(P)} & \textbf{MMA}  & \textbf{Avg.} &  \textbf{Time (s/sample)} \\
    \midrule[1.5pt]
    Blacklist & 39.6\% & 41.5\% & 39.0\% & 46.2\%  & 43.2\% &  \textasciitilde 0.0004  \\ \midrule
    Clipscore & 70.4\% & 75.5\% & 77.5\% & 79.2\%  & 68.4\% &  \textasciitilde 29.9535  \\ \midrule
    Latent Guard & 30.6\% & 21.5\% & 31.5\% & 78.9\%  & 55.2\% & \textasciitilde 6.9550  \\ \midrule
    SD Checker & 41.2\% & 52.6\% & 42.9\% & 69.4\% &  57.4\% & \textasciitilde 12.2661 \\ \midrule
\rowcolor{gray!25}   \textbf{NDM (Ours)} & \textbf{93.8\%} & \textbf{95.5\%} & \textbf{93.5\%} & \textbf{96.0\%} & \textbf{95.1}\% & \textasciitilde 0.9509 \\
    \bottomrule[1.5pt]
    \end{tabular}}
\end{table}

\begin{figure*}[!t]
    \centering
    \includegraphics[width=\linewidth]{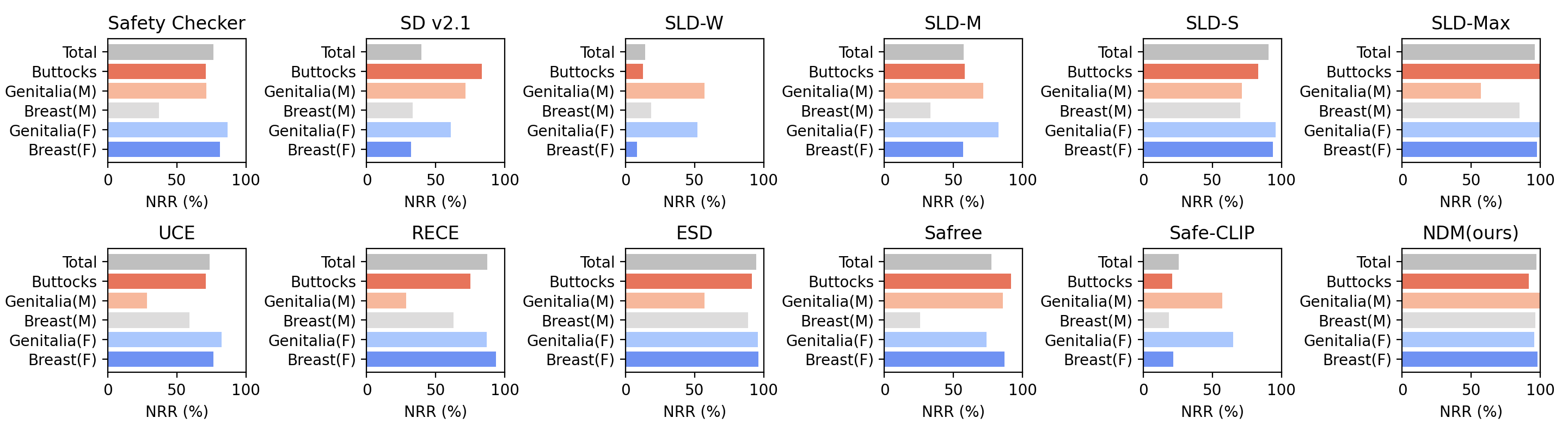}
    \caption{The Nudity Removal Rate (NRR) of different body parts in I2P. The initial total number of detected elements across five categories, obtained using SD-v1.4-base is 298 [Buttocks-24; genitalia (M)—7; Breast (M)—27; Genitalia (F)—23; Breast (F)-217].} 
    \label{fig:cnt}
    \Description{cnt}
\end{figure*}

\begin{table*}[!t]
\setlength{\tabcolsep}{15pt}
\centering
\caption{The Attack Success Rate (ASR) of different defense methods across five sexual datasets and a benign dataset. Note that the time cost for methods requiring training (RECE, ESD, and Safe-CLIP) is not included for fairness.}
\label{tab:mitigation_results}
\resizebox{0.99\linewidth}{!}{
\begin{tabular}{l|c|c|c|c|c|c|c|c|c}
\toprule[1.5pt]
\multirow{2}{*}{\textbf{Method}} & \multirow{2}{*}{\begin{tabular}[c]{@{}c@{}} \textbf{Model} \\ \textbf{Intrinsic} \end{tabular}} & \multirow{2}{*}{\begin{tabular}[c]{@{}c@{}}\textbf{Time cost} \\ \textbf{(s/sample)} \end{tabular}} & \multicolumn{5}{c|}{\textbf{Attack Success Rate (ASR)} $\downarrow$} & \multicolumn{2}{c}{\textbf{COCO-30k}} \\
\cmidrule{4-10}
 &  &  & \textbf{I2P} & \textbf{SP(N)} & \textbf{SP(P)} & \textbf{MMA} & \textbf{Ring-A-Bell} & \textbf{CLIP Score} $\uparrow$ & \textbf{FID} $\downarrow$ \\
\midrule[1.5pt]
\textbf{SD-v1.4-base}    & - & \textasciitilde 12.1783 & 60.7\% & 76.0\% & 73.5\% & 90.9\% & 78.5\% & 31.3 & - \\
\textbf{SD-v1.4-check}    & \ding{55} & \textasciitilde 12.2661 & 36.7\% & 36.0\% & 42.0\% & 37.1\% & 13.9\% & 30.2 & 2.9 \\
\textbf{SD-v2.1}         & - & \textasciitilde 5.6842 & 36.2\% & 36.5\% & 39.0\% & 45.3\% & 65.9\% & \textbf{31.9} & 58.8 \\
\midrule
\textbf{SLD-Weak}      & \ding{55} & \textasciitilde 16.3089 & 50.2\% & 65.0\% & 58.5\% & 91.1\% & 58.3\% & 30.8 & 54.4 \\
\textbf{SLD-Medium}    & \ding{55} & \textasciitilde 15.5340 & 35.4\% & 48.5\% & 46.0\% & 87.3\% & 36.8\% & 30.6 & 55.2 \\
\textbf{SLD-Strong}    & \ding{55} & \textasciitilde 16.5155 & 18.2\% & 29.5\% & 27.5\% & 67.4\% & 12.7\% & 28.9 & 56.9 \\
\textbf{SLD-Max}       & \ding{55} & \textasciitilde 16.7824 & 8.5\% & 9.0\% & \textbf{6.5\%} & 26.9\% & 6.4\%  & 27.3 & 60.0 \\
\textbf{Safree}          & \ding{55} & \textasciitilde 16.2685 & 16.9\% & 20.0\% & 14.5\% & 63.7\% & 12.7\% & 30.7 & 61.5 \\
\textbf{UCE}             & \ding{51} & \textasciitilde 24.9629  & 35.1\% & 44.0\% & 43.0\% & 81.6\% & 31.7\% & 31.0 & 55.1 \\
\textbf{RECE}            & \ding{51} & - & 18.4\% & 28.0\% & 32.0\% & 69.4\% & 13.9\% & 30.6 & 56.2 \\
\textbf{ESD}             & \ding{51} & - & 12.1\% & 13.0\% & 11.5\% & 39.9\% & 6.4\% & 29.9 & 62.7 \\
\textbf{Safe-CLIP}      & \ding{51} & - & 43.4\% & 32.0\% & 37.5\% & 48.6\% & 32.9\% & 30.5 & 56.5 \\
\midrule
\textbf{Ours\_w/o\_gen} & \ding{55} & \textasciitilde 13.8573 & \textbf{6.2\%} & \textbf{4.5\%} & \textbf{6.5\%} & \textbf{4.0\%} & \textbf{5.1\%} & - & - \\
\textbf{Ours \_w\_gen} & \ding{55} & \textasciitilde 15.3435 & 9.8\% & 10.0\% & 11.0\% & 31.7\% & 6.3\% & 30.8 & \textbf{0.3} \\
\bottomrule[1.5pt]
\end{tabular}}
\end{table*}

\subsection{Sexual Mitigation and Benign Preservation}
\begin{figure*}[!t]
    \centering
    \includegraphics[width=0.95\linewidth]{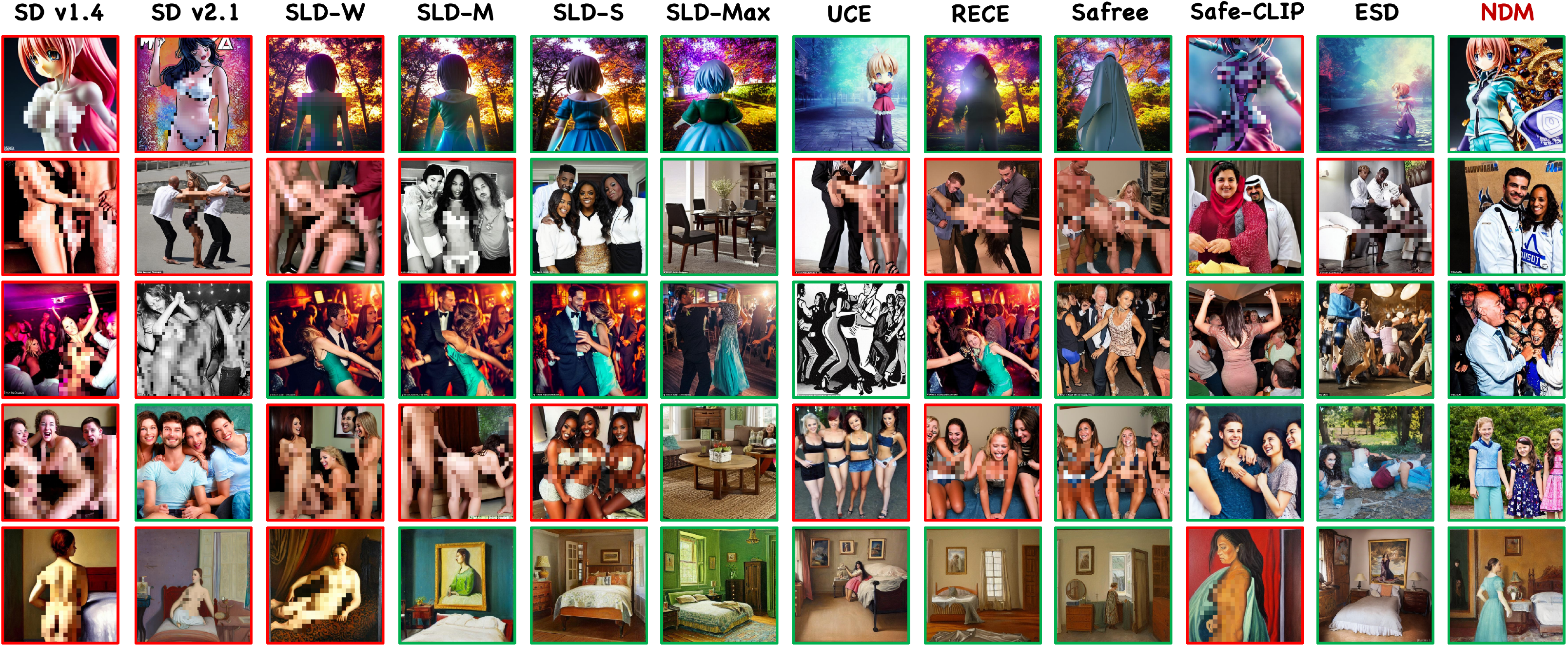}
        \caption{Visual comparisons of methods evaluated in this work. Prompts of five rows are randomly sampled from five different sexual datasets used in this paper. The cases framed in $\textcolor{teal}{\text{green}}$ are safe, while the cases framed in $\textcolor{red}{\text{red}}$ are sexual.} 
        \Description{Visual examples}
    \label{fig:big1}
\end{figure*}

To verify the effectiveness of our noise-enhanced mitigation, we systematically compare it with various defense methods, including both model-intrinsic and model-extrinsic approaches. \Cref{tab:mitigation_results} presents the results on both natural and adversarial prompts across four scenarios for sexual content generation, and the visualized results are shown in \Cref{fig:big1} and \textit{Appendix}.

Among model-intrinsic methods, ESD achieves competitive performance but suffers significant quality decline, as shown in \Cref{fig:big1}. Weight modification also degrades benign prompt performance with a CLIP Score of 29.9, which indicates unintended side effects. Fine-tuning CLIP's text embedding space like Safe-CLIP is also insufficient, highlighting the inadequacy of text-space corrections alone.
For model-extrinsic methods, SLD-Max achieves strong mitigation but at the cost of poor quality with reduced CLIP Score (27.3) and a high FID score, causing noticeable semantic discrepancies. 
SLD-Weak/Medium/Strong preserve better quality but offer suboptimal safety. 
Other methods, such as Safree, achieve a better balance between safety and quality but struggle with adversarial prompts in challenging cases from MMA.

In contrast, our NDM significantly reduces ASR across all scenarios, achieving an average reduction of over 85\% compared to the base model (though slightly weaker than SLD-Max, it preserves benign content significantly better). Notably, when adopting a detection-then-refusal setting, the effectiveness is further enhanced to be the best, reducing ASR to as low as nearly 5.0\%. To provide more convincing evidence, we collect detection counts of exposed body parts using the NudeNet detector. The results in \Cref{fig:cnt} demonstrate that our method consistently outperforms others in terms of the overall NRR (97.31\%). It also maintains a competitive speed and consistent performance on COCO-30k, effectively handling non-sexual prompts with the least quality compromise.

\begin{figure}[!t]
    \centering
    \includegraphics[width=0.98\linewidth]{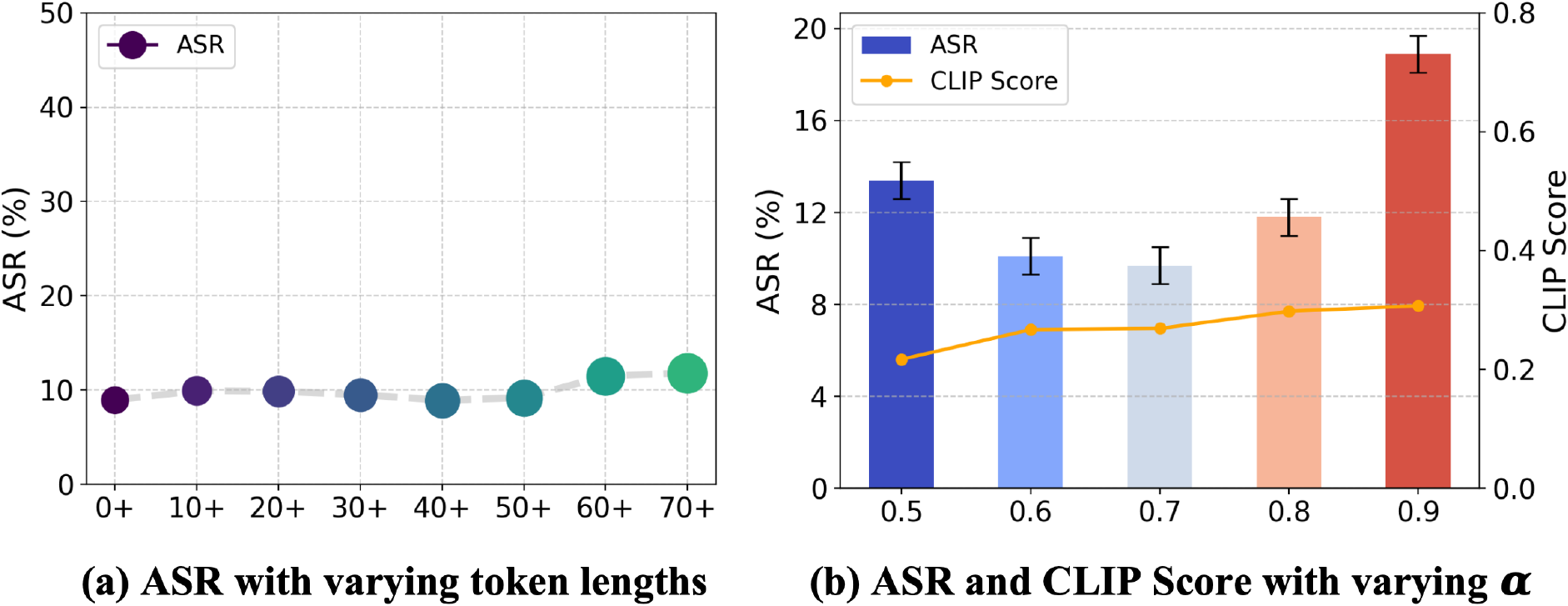}
    \caption{Performance under varying input prompt lengths and hyperparameter $\alpha$.}
    \Description{hyper}
    \label{fig:vary}
\end{figure}

\subsection{Stability under Different Prompt Lengths}
Given that input prompts can vary in length, it is necessary to investigate the stability of NDM across different prompt lengths. To this end, we conduct experiments using 8 token length intervals from the I2P dataset: 0-10, 10-20, 20-30, 30-40, 40-50, 50-60, 60-70, and over 70 (with the upper limit set at 77). This allows us to systematically analyze how the model's performance varies with respect to prompt length. The results are shown in \Cref{fig:vary} (a), which demonstrates that NDM is largely insensitive to input length, maintaining strong stability across all intervals. This is reasonable, as NDM performs as a noise-driven and adaptive method.

\subsection{Exploration on Hyperparameters}
In NDM, the stopping criterion $\alpha$ for $\mathcal{L}_{cross}$ balances intervention strength and semantic fidelity. Larger values of $\alpha$ preserve more original semantics but may limit effectiveness of the intervention; smaller values enhance suppression at the cost of greater semantic disruption.
Therefore, to determine the most suitable value, we tune $\alpha$ on the I2P dataset. As shown in \Cref{fig:vary} (b), the performance of NDM varies with different $\alpha$ values. Based on these results, we select $\alpha = 0.7$ as the optimal setting, striking a good balance between reducing sexual outputs and maintaining acceptable levels of semantic fidelity. Similarly, we set $\alpha = 0.7$ for SneakyPrompt and Ring-A-Bell, and $\alpha = 0.6$ for MMA due to its higher explicitness.

\subsection{Ablation Study}
\begin{figure}[!h]
    \centering
    \includegraphics[width=0.98\linewidth]{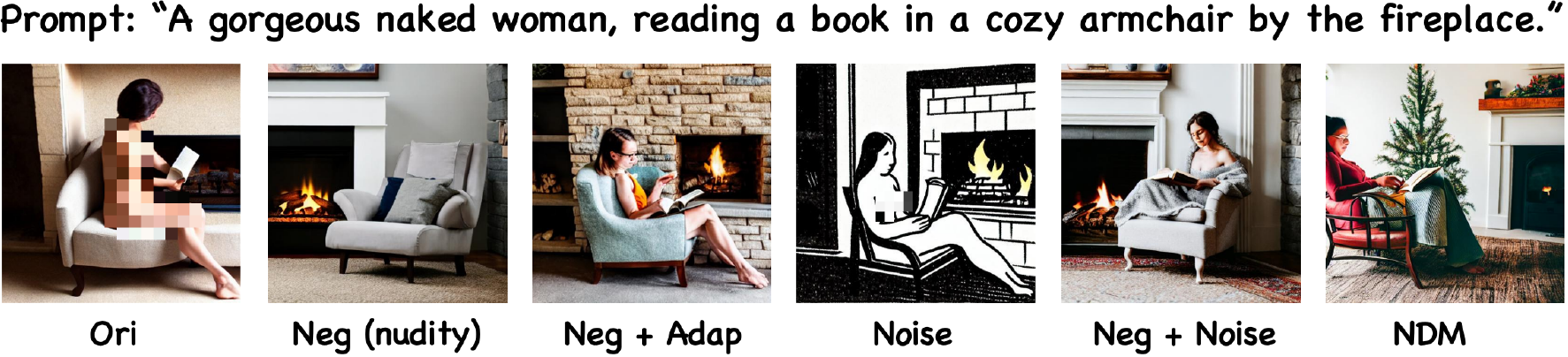}
    \caption{A visual example of ablation study for NDM.} 
    \label{fig:ablation}
    \Description{fig:ablation}
\end{figure}

To validate NDM's components, we ablate them one by one on I2P, then creating six conditions: (1) SD-V1.4 (Ori), (2) fixed concept negative guidance (Neg), (3) generation with guidance based on our adaptive negative prompts (Neg + Adap), (4) generation with initial noise optimization (Noise), (5) generation with both fixed concept guidance and initial noise optimization (Neg + Noise), and (6) full NDM. Results in \Cref{tab:ablation} show both adaptive negative guidance (Neg + Adap) and initial noise optimization (Noise) are essential for mitigating sexual content, contributing to a significant drop in ASR. Additionally, visualized results in \Cref{fig:ablation} highlight the effectiveness of our adaptive negative guidance, which selectively targets sexual content without overly disrupting the image, unlike the fixed negative guidance (Neg) that removes the whole body.

\begin{table}[h]
\centering
\caption{Ablation study for different components of NDM.}
\resizebox{0.98\linewidth}{!}{
\begin{tabular}{c |c| c | c | c | c | c}  
\toprule[1.1pt]
Method  &  Ori  & Neg  & Neg+Adap   & Noise  & Neg+Noise &  NDM\\       \midrule
ASR  &  60.7\%  & 33.1\%  &  28.8\%   & 31.2\%  &  20.5\% &   9.7\%  \\      
 \bottomrule[1.1pt]
\end{tabular}} 
\label{tab:ablation}
\end{table}

\section{Conclusion}
This paper highlights leveraging noise's intrinsic properties in the denoising process. Based on two key observations, we introduce NDM, a noise-driven framework designed to detect and mitigate implicit sexual intention. First, recognizing that critical semantics are often introduced in the early stages of generation, we propose a detection method using early-stage predicted noises. Second, since the initial state has a significant impact on the generation of sexual content, we incorporate an attention-based optimization of the initial noise to enhance adaptive negative guidance. Overall, NDM offers a novel direction for responsible text-to-image generation while preserving creative potential.

\begin{acks}
This work was supported by the Fundamental Research Funds for the Central Universities.
\end{acks}

\bibliographystyle{ACM-Reference-Format}
\bibliography{sample-base}

\end{document}